\pdfoutput=1

\documentclass[11pt]{article}

\usepackage{comment}

\usepackage{acl}
\usepackage{color,soul}
\usepackage{amssymb}
\usepackage{xcolor}
\usepackage{times}

\usepackage{latexsym}
\usepackage{algorithm}
\usepackage{algpseudocode}
\usepackage{tabularx}
\usepackage{hyperref}

\usepackage{subcaption}
\usepackage{mwe}

\usepackage{mathtools}
\usepackage{lipsum}
\usepackage{float}

\usepackage[T1]{fontenc}

\usepackage[utf8]{inputenc}

\usepackage{microtype}

\title{Partner Personas Generation for Diverse Dialogue Generation}

\author{Hongyuan Lu, Wai Lam, Hong Cheng, Helen M. Meng \\
  The Chinese University of Hong Kong \\
  \texttt{\{hylu,wlam,hcheng,hmmeng\}@se.cuhk.edu.hk} \\}

\begin{document}
\maketitle
\begin{abstract}
Incorporating personas information allows diverse and engaging responses in dialogue response generation. Unfortunately, prior works have primarily focused on self personas and have overlooked the value of partner personas. Moreover, in practical applications, the availability of ground truth partner personas is often not the case. This paper attempts to tackle these issues by offering a novel framework that leverages automatic partner personas generation to enhance the succeeding dialogue generation. We incorporate reinforcement learning with a dedicatedly designed critic network for reward judgement. Experimental results from both automatic and human evaluation demonstrate: \textbf{a)} Our framework is capable of generating relevant, informative and coherent partner personas, even compared to the ground truth partner personas. \textbf{b)} Generated partner personas enhance the succeeding response generation, thus surpassing our baselines and comparison model \textit{when partner personas are missing during the inference stage}. \textbf{c)} Our framework generates responses that are more informative and engaging than our baseline \textit{conditioned on the ground truth partner personas during inference}. \textbf{d)} Our dedicatedly designed critic network reinforces our framework effectively. Finally, our framework gives better explainability and reduces the demands for external databases for partner personas.\footnote{We will release our data and our code upon publication.}
\end{abstract}
\section{Introduction}
Building informative and engaging dialogue agents \citep{2019arXiv191100536Z, 2020arXiv200413637R, 2021arXiv210909519B} is a popular research direction within the area of natural language processing. For the sake of engagement, diverse and consistent responses \citep{song-etal-2020-generate, song-etal-2021-bob} are important factors, and personas information \citep{PERSONACHAT} gives rise to both of the factors. There are two types of personas information, namely self persona and partner persona. The former refers to a self profile consisting of several sentences representing the dialogue agent. Such a persona allows the agent to produce consistent responses rather than solely relying on the personas information \citep{kim-etal-2020-will} that is randomly learned and embedded in the model parameters. The latter also refers to a profile but representing the user. Leveraging such partner personas is helpful for dialogue generation \citep{PTMT}. Therefore, we exploit partner personas for diverse dialogue generation. 
\par Unfortunately, the user profile could be commonly missing due to the cold-start \citep{2020arXiv200512979L} when deploying online dialogue agents or for newly registered users. Most of the works, if not all, \citep{FTPC, 2019arXiv190512188S, DIM, DGMN} have been either overlooking the value of partner personas or simply focusing on the impractical situation where partner personas guarantee to exist in both training and inference stages. Our work demonstrates the importance of diverse partner personas generation, and we particularly investigate the practical situation when partner personas are missing in the inference stage. Such an investigation is essential as there is no guarantee that the ground truth partner personas always exist. \textit{Ultimately, our proposed framework can produce even more diverse and engaging responses than our baseline that conditions on the ground truth partner personas.} We demonstrate a case study in Section \ref{case} that illustrates this advantage. 
\par To our best knowledge, this is the first attempt to formulate partner personas prediction in a generative manner that boosts the performance of the succeeding dialogue generation. Our work is motivated by three underlying hypotheses: \textbf{i)} Partner personas generation is plausible given the self personas and dialogue context. \textbf{ii)} Generated personas are more diverse and interesting than the retrieved ground truth. \textbf{iii)} Such diverse generated personas help to produce diverse succeeding dialogue responses. Our automatic and human evaluation results support these hypotheses, and this paper paves the way to exploit generative partner personas for diverse dialogue generation.
\par
We develop a novel framework composed of three major components, namely a partner personas generator, a dialogue response generator and a critic network. We use a partner personas generator to generate partner personas, which the dialogue response generator uses for succeeding dialogue response generation. We employ reinforcement learning with a dedicatedly designed critic network that propagates the reward back to the generators.
\par
Prior works have investigated retrieval-based partner persona \citep{PERSONACHAT,2019arXiv190512188S}. The human-constructed ground truth personas serve as the upper bound for such retrieval-based systems, and we argue that the ground truth are not diverse enough. We observe that the generative counterpart generates relevant, informative and coherent partner personas, which further diversifies the succeeding dialogue response generation. It follows another advantage that our framework does not need an external database to retrieve from \citep{madotto-etal-2020-learning, free}. 
\par
One close work to ours is a multi-task framework for meta-learning \citep{MTML} that uses personas reconstruction as an auxiliary task to improve response consistency. The differences are that theirs does not differentiate between self personas and partner personas, while ours does. Their model focuses on meta-learning, while ours does not set such a constraint. Theirs indicates an improvement over personality consistency only, while ours focuses on diverse dialogue generation. We conduct an empirical comparison with their model, reconstructing the partner personas. Experimental results indicate that such a multi-task model does not work well in our problem setting.
\par
We compare our proposed framework with some competitive baselines. The automatic and human evaluation results indicate that our framework can generate even more diverse and engaging responses than the baseline conditioned on ground truth partner personas. It leads to a conclusion that \textbf{i)} Partner personas generation is plausible. \textbf{ii)} The generated partner personas are more diverse than the ground truth partner personas. \textbf{iii)} Our framework produces even more diverse and engaging responses than our competitive baselines that condition on the ground truth partner personas.
\section{Related Work}
\subsection{Personalized Dialgoue Generation}
Conditioning on personas helps produce informative and engaging responses. The largest multi-turn dialogue dataset conditioned on personal profiles is \textsc{PersonaChat}, in which two crowdsourcers converse and find more about each other. To better utilise the self personas in generating consistent responses, the community has proposed quite a lot of methods. \citet{FTPC} employs a pre-training stage based on dedicatedly extracted large-scale persona-based dialogues and fine-tunes the model on \textsc{PersonaChat}. \citet{DGMN} fuses information in personas and dialogue context into individual contextualised representations by attending to different parts of both. \citet{DIM} exploits the interaction between personas, dialogue context and response to improve retrieval-based dialogue agents. \citet{MTML} utilises multi-task learning for improved personality consistency in the meta-learning scenario. \citet{PTMT} employs four different strategies for personas fusing, which learns to use self persona and partner persona in a more effective manner. There have also been several works based on GPT \citep{2019arXiv190108149W}. However, most of these prior works focus on exploiting self personas rather than partner personas, and they have been assuming that the ground truth partner personas guarantee to exist.
\subsection{Reinforcement Learning}
\label{critic}
Reinforcement learning (RL), or specifically, policy gradient methods \citep{policygradients}, have been frequently adopted to both task-oriented dialogue agents \citep{RMM, 2021arXiv210509710D} or open-domain chitchat agents \citep{RLChat, 2019arXiv190907547S}. It can either propagate non-differentiable loss \citep{cai-etal-2019-skeleton} or optimize an expert reward such as ease of answering \citep{RLChat}. It also adopts a scenario where a user simulator and a dialogue agent interact, and an expert reward function is defined to assign the goodness to each response generated \citep{RMM}. 
\begin{figure}[t!]
\begin{center}
\vspace{0mm}
\centerline{
\includegraphics[width=6cm]{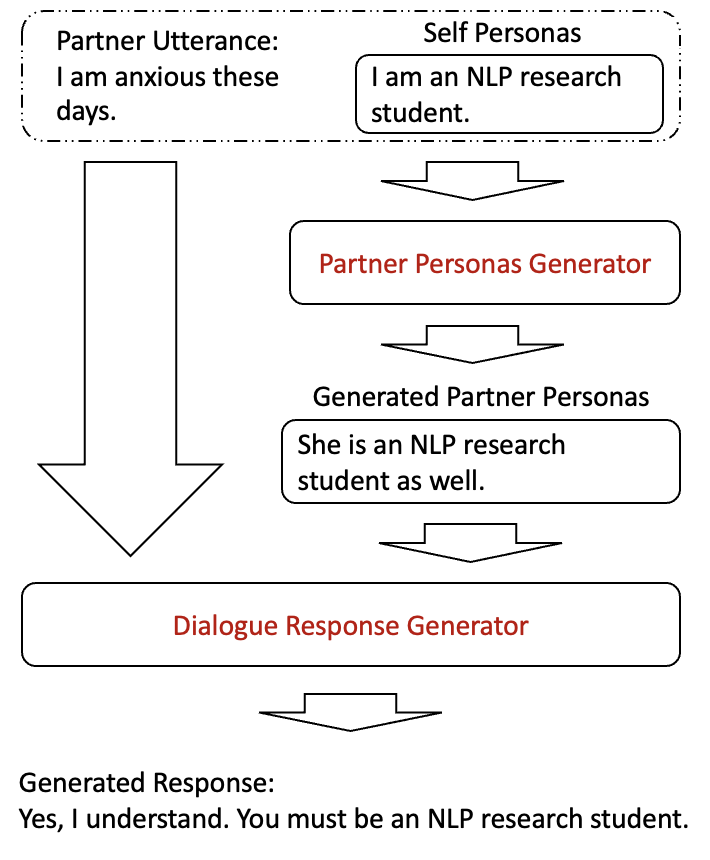}}
\caption{An example of the inference flow that shows the generated partner personas and the incorporation of partner personas generation into response generation.}
\label{2}
\end{center}
\end{figure}
\begin{figure}[t!]
\begin{center}
\vspace{0mm}
\centerline{
\hspace{-5mm}
\includegraphics[width=8.5cm]{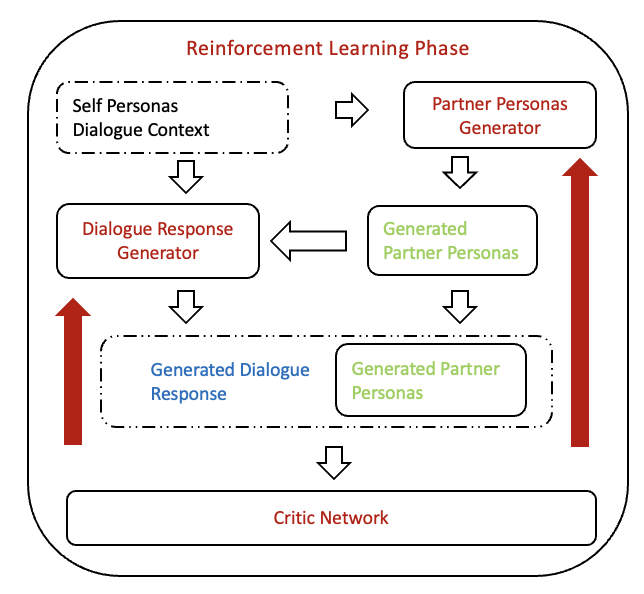}}
\caption{The illustrated reinforcement learning strategy that directly backpropagates the response-related rewards from the critic network to the partner personas generator and the dialogue response generator. }
\label{1}
\end{center}
\end{figure}
\section{Proposed Framework}
We develop a novel framework composed of three major components, namely a partner personas generator, a dialogue response generator and a critic network used by reinforcement learning. Figure \ref{2} depicts the inference flow of our setting. The input dialogue context with self persona is first fed into the partner personas generator. The generated partner personas output is then concatenated with the dialogue context and the self personas as the input into the dialogue response generator. In the beginning, we train our partner personas generator and dialogue response generator \textbf{separately} under supervised learning. In the training stage, we use the ground truth partner personas to train the dialogue response generator, and we replace it with generated partner personas in the inference stage. After the supervised learning stage, the second stage is a reinforcement learning stage which \textbf{jointly} optimizes both partner personas generator and dialogue response generator as depicted in Figure \ref{1}. Such framework has two advantages: \textbf{i)} The partner personas generator can be trained directly using the reward signal that is relevant to dialogue response generation. \textbf{ii)} The dialogue response generator trained using ground truth partner personas can be further fine-tuned on the generated partner personas.\footnote{Section \ref{ablation} presents an ablation study on reinforcement learning that supports this claim.} Particularly, we employ a dedicatedly designed critic network that receives generated partner personas and generated dialogue responses as the input and output a reward that measures the relevance between the generated personas and responses and propagates back to the generators.
\subsection{Partner Personas Generation}
A Seq2Seq neural network \citep{2014arXiv1409.3215S} is adopted as our partner personas generator for the task of partner personas generation (PPG). The concatenation of dialogue context \textbf{c} and self personas $\mathbf{s}$ is fed as an input into the partner personas generator. The personas generator then outputs an approximated partner personas $\mathbf{\hat{p}}$ conditioned on the input, which maximises the following conditional likelihood: \begin{align*}
    P(\mathbf{\hat{p}}\mid \mathbf{s}, \mathbf{c})=\prod_{t=1}^{T}P(\hat{p}_t\mid \hat{p}_1,..., \hat{p}_{t-1}, \mathbf{s}, \mathbf{c})
\end{align*}
For training, the ground truth partner personas $\mathbf{p}$ is used and we train our generator under the maximum likelihood estimation: 
\begin{align*}
    P(\mathbf{p}\mid \mathbf{s}, \mathbf{c})=\prod_{t=1}^{T}P(p_t\mid p_1,..., p_{t-1}, \mathbf{s}, \mathbf{c})
\end{align*}

\subsection{Dialogue Response Generation}
We also adopt a Seq2Seq neural network for the task of dialogue response generation (DRG). The concatenation of dialogue context \textbf{c}, self personas $\mathbf{s}$, and partner personas $\mathbf{\hat{p}}$ is fed as an input into the dialogue response generator. The personas generator then outputs an approximated dialogue response $\mathbf{\hat{r}}$ conditioned on the input, which maximises the following conditional likelihood:
\begin{align*}
    P(\mathbf{\hat{r}}\mid \mathbf{s}, \mathbf{\hat{p}}, \mathbf{c})=\prod_{t=1}^{T}P(\hat{r}_t\mid \hat{r}_1,..., \hat{r}_{t-1}, \mathbf{s}, \mathbf{\hat{p}}, \mathbf{c})
\end{align*}
For training, the ground truth partner personas $\mathbf{p}$ and the ground truth dialogue response $\mathbf{r}$ is used and  we train our generator under the maximum likelihood estimation: 
\begin{align*}
    P(\mathbf{r}\mid \mathbf{s}, \mathbf{p}, \mathbf{c})=\prod_{t=1}^{T}P(r_t\mid r_1,..., r_{t-1}, \mathbf{s}, \mathbf{p}, \mathbf{c})
\end{align*}
We use the ground truth partner personas for training and generated partner personas for inference.

\subsection{Reinforcement Learning}
\label{rll}
We employ a critic network that is the core of our reinforcement learning (RL) algorithm to reward our reinforcement agents. We train a binary classifier as our critic network by extracting sub-training-instances $(\mathbf{s}, \mathbf{r}, L=1)$, where $L=1$ represents positive training samples. We then randomly sample two distinct positive sub-instances $A$ and $B$:
\[(\mathbf{s}^{A}, \mathbf{r}^{A}, L=1) \]
\[(\mathbf{s}^{B}, \mathbf{r}^{B}, L=1) \]
Then two negative samples can be derived as:
\[(\mathbf{s}^{A}, \mathbf{r}^{B}, L=0) \]
\[(\mathbf{s}^{B}, \mathbf{r}^{A}, L=0) \]
Thereafter, we fine-tune a binary classifier as our critic on this training set by minimizing the following binary cross-entropy loss:
\[-Llog(P(L\mid \mathbf{s}, \mathbf{r}))-(1-L)log(1-P(L\mid \mathbf{s},\mathbf{r}))\]\\
In the equation above, the binary label $L$ indicates \textit{whether the given response is relevant to the given personas}. During the reinforcement learning stage, this classifier acts as a critic network that outputs $\hat{L}$, conditioned on the generated partner personas $\mathbf{\hat{p}}$ and generated response $\mathbf{\hat{r}}$. The predicted binary label $\hat{L}$ is then converted to a reward $R$. $R$ is a positive reward when $\hat{L}=1$, and $R$ is a negative reward when $\hat{L}=0$.
We then update our RL agents with the following gradients:
\[\Delta \theta_{PPG} = R\triangledown_{\theta_{PPG}} -log P(\mathbf{\hat{p}}\mid \mathbf{s},\mathbf{c}) \]
for the partner personas generator (PPG), and for the dialogue response generator (DRG):
\[\Delta \theta_{DRG} = R\triangledown_{\theta_{DRG}}-log P(\mathbf{\hat{r}}\mid \mathbf{s},\mathbf{\hat{p}},\mathbf{c}) \]
We particularly want to give positive rewards to our RL agents when they give high-quality responses that differ from the \textit{ground truth}. Since it is not straightforward to understand the underlying motivation for such a critic network, we divide it into two cases and conquer each of them:
\begin{itemize}
    \item The critic network outputs $0$ as $\hat{L}$: the generated personas and response are irrelevant, and we assign a negative reward.
    \item The critic network outputs $1$ as $\hat{L}$: the generated personas and response are relevant, and we assign a positive reward.
\end{itemize}
The first case is trivial, as it is reasonable to assign a negative reward when at least one of our RL agents generates an output far away from the ground truth. For the second case, in addition to the trivial case that both agents output ground-truth-like generation, it also considers such a case when both partner personas generator and dialogue response generator generate a relevant output, but not the exact ground truth. Maximum likelihood estimation might fail to capture this reward, as there could still be a certain distance to the ground truth. In contrast, our critic network captures this by outputting $1$ and assigns both of our RL agents a positive reward.
We design such a dedicate reward mechanism to encourage the generator to produce a diverse and engaging response with the diverse partner personas generated. We present a case study in Section \ref{case} that illustrates this advantage.
\par
Previous work \citep{cai-etal-2019-skeleton} employed critic network for reinforcement loss backpropagation. At first glance, our usage of the critic network shares some resemblances to theirs, but indeed, the underlying motivation vastly differs.
The major difference is that their critic is trained in an adversarial manner \citep{2018arXiv180400861L} to pick up the \textit{ground truth} response among other negative candidates. Also, their critic network conditions only on the dialogue response but not on the generated skeleton. In contrast, we further diversify the response generation with a classifier conditioning on both the generated personas and the generated response.

\begin{table*}
\small
\centering
\begin{tabular}{lccccc}
\hline
\noalign{\vskip 1mm}  
\textbf{Model} & \textbf{PPL$\downarrow$} & \textbf{ROUGE} & \textbf{METEOR} & \textbf{Distinct-1}& \textbf{Distinct-2}\\
\noalign{\vskip 1mm}  
\hline
\hline
\noalign{\vskip 1mm}  
E2E w/o Partner Personas & $17.86$ & $17.21$ & $24.62$ & $0.01278$  & $0.1502$ \\
E2E w/ Training Partner Personas & $17.45$ & $16.91$ & $24.29$ & $0.01278$  & $0.1485$ \\
Multi-task Learning \citep{MTML} & $177.5$ & $10.33$ & $12.85$ & $0.00423$  & $0.0574$ \\
\noalign{\vskip 1mm}  
\hline
\hline
\noalign{\vskip 1mm}  
Our Framework w/ RL PPG & $17.01$ & $17.15$ & $25.55$ & $0.01304$  & $0.1602$ \\
Our Framework w/ RL DRG & $\mathbf{16.87}$ & $17.46$ & $25.88$ & $0.01450$  & $0.1664$ \\
\noalign{\vskip 1mm} 
\hline
\noalign{\vskip 1mm} 
Our Framework w/o RL & $17.39$ & $17.43$ & $24.81$ & $0.01344$  & $0.1543$ \\
\textbf{Our Framework w/ RL PPG\&DRG} & $17.16$ &  $\mathbf{17.51}$ &  $\mathbf{26.36}$ &  $\mathbf{0.01495}$  &  $\mathbf{0.1726}$ \\ 
\noalign{\vskip 1mm}  
\hline
\hline
\end{tabular}
\caption{\label{main_result}
Test results on the \textsc{PersonaChat} dataset. Perplexity (PPL) attains better quality with lower scores, and the remaining metrics attain better quality with higher scores. PPG represents partner personas generation and DRG represents dialogue response generation. We leave corresponding validation performance in Appendix \ref{valid}. \
}
\end{table*}

\section{Experimental Setup}
\subsection{Dataset}
We conduct all our experiments on \textsc{PersonaChat} \citep{PERSONACHAT}, the largest multi-turn dialogue dataset conditioned on personas profile. We follow the training/validation/testing split from the ParlAI platform \citep{2017arXiv170506476M}, which contains about 65,000 training instances, about 7,800 validation instances and about 7,500 testing instances. As for the reinforcement learning in Section \ref{rll}, we collect about 130,000 training instances from the training partition with equally distributed positive and negative samples to train our critic network.
\begin{figure}[t!]
\begin{center}
\vspace{0mm}
\centerline{
\includegraphics[width=8cm]{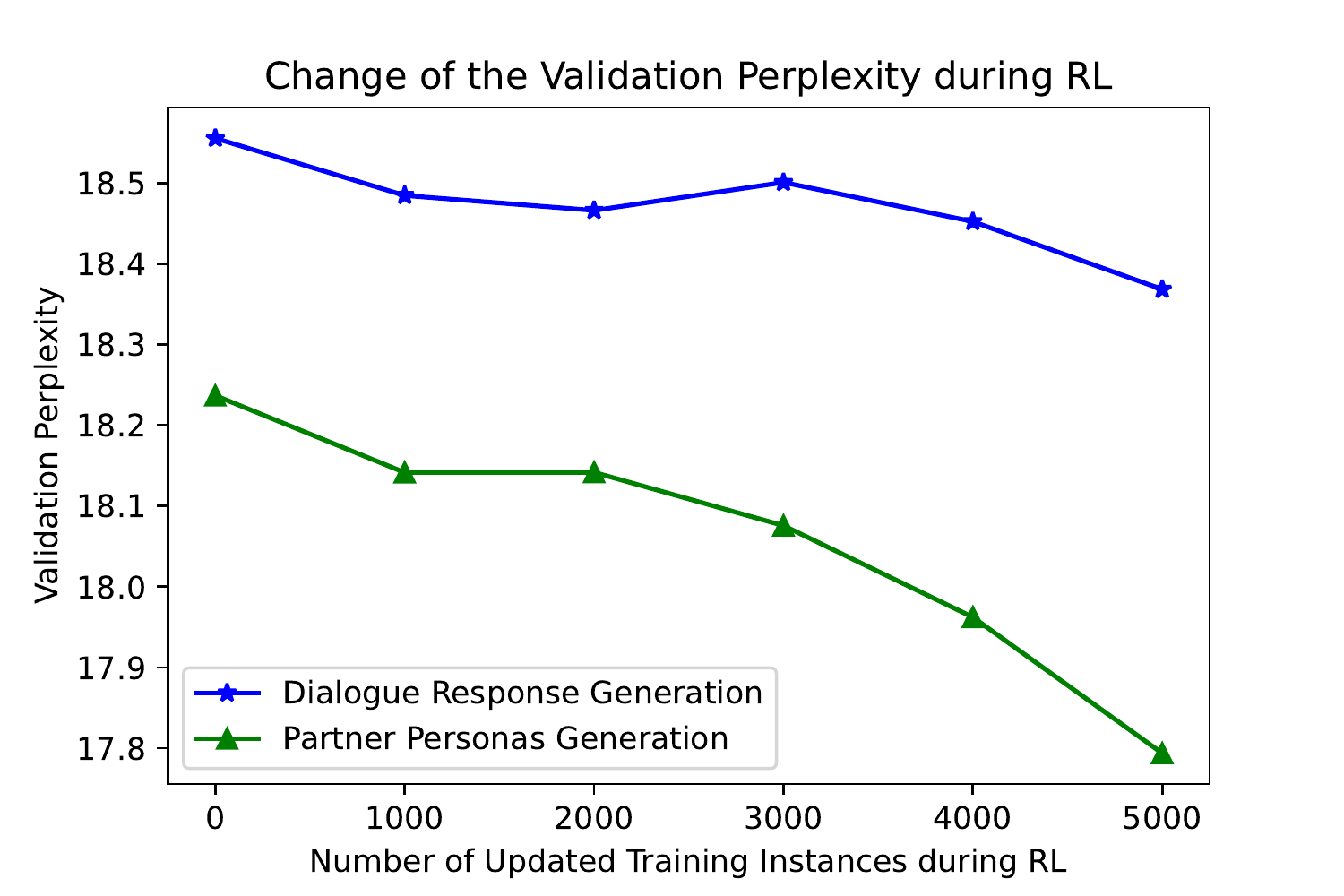}}
\caption{Change of the validation perplexity for our proposed framework during RL.}
\label{3}
\end{center}
\end{figure}
\subsection{Baselines and Comparison Models}
\paragraph{End-to-end Baseline without Partner Personas}  Our first baseline is an end-to-end response generator without any partner persona information throughout the experiment. We notice that it could be unfair to directly compare our proposed framework with this baseline, as our framework uses ground truth partner personas in the training stage for our proposed framework. Since it does not use the same amount of training information as our proposed framework, we offer our second baseline trained with the ground truth partner personas.

\paragraph{End-to-end Baseline with Training Partner Personas} Our second baseline is a an end-to-end response generator that is trained using ground truth partner personas. In the inference stage, we feed the concatenation of self personas and dialogue context as the input. For the sake of fairness, \textit{it uses the same amount of training information and inference information as our proposed framework.}

\paragraph{Multi-task Learning Comparison Model}
Following prior work \citep{MTML}, we build a multi-task learning comparison model with partner personas generation as an auxiliary task. The model is trained to maximise the training objective of the sum of the partner personas generation labels likelihood $L_{PPG}$, and the dialogue response generation labels likelihood $L_{DRG}$. Both of the tasks are generated and conditioned on dialogue context and self personas by sharing the same model parameters. We maximise the loss:
\begin{align*}
    \alpha L_{PPG} + L_{DRG},
\end{align*}
where $\alpha$ is a loss weighting parameter which we tune it over the validation set.

\subsection{Implementation Details}
For all of our baselines, comparison model, the partner personas generator and the dialogue response generator, we use pre-trained GPT-2 \citep{gpt2} to initialize their model parameters. For the supervised phase, we set Adam \citep{adam} as our optimizer, with hyperparameters $\eta=5e{-}4$, $\beta_1=0.9$, $\beta_2=0.999$, $\epsilon=1e{-}8$. We fine-tune 2 epochs on all the baselines, comparison models and our proposed framework modules and select the models with the lowest validating perplexity. For the RL phase, we set Adam as our optimizer, with hyperparameters $\eta=5e{-}6$, $\beta_1=0.9$, $\beta_2=0.999$, $\epsilon=1e{-}8$. We update the model parameters every $20$ training instances and validate the model performance every $50$ updates. For our critic network for reward judgement in the RL phase, we use DistilBERT \citep{2019arXiv191001108S} to initialize the model parameters. We set Adam as our optimizer, with hyperparameters $\eta=5e{-}6$, $\beta_1=0.9$, $\beta_2=0.999$, $\epsilon=1e{-}8$. We fine-tune the critic for 1 epoch on the original training split from the \textsc{PersonaChat}. We conduct all our experiments based on the \textit{Transformers} library from Huggingface \citep{wolf-etal-2020-transformers}.
\subsection{Evaluation Metrics}
We report intrinsic perplexity to evaluate the model with the ground truth response \citep{2020arXiv200413637R}. We report distinct-1 and distinct-2 \citep{2015arXiv151003055L}  to evaluate model diversity, which calculate the ratio of distinct unigrams/bigrams against total unigrams/bigrams generated. We report ROUGE \citep{rouge} and METEOR \citep{banerjee-lavie-2005-meteor2} as the extrinsic evaluations.
\section{Results}
Recall our four claims: \textbf{a)} Our framework generates diverse partner personas than the ground truth partner personas. \textbf{b)} Partner personas generation benefits the succeeding response generation, thus surpassing our baselines. \textbf{c)} Our framework generates more diverse and engaging responses than our competitive baseline that uses ground truth training and inference partner personas. \textbf{d)} We employ reinforcement learning with a dedicatedly designed critic network that effectively boosts our framework.
Also, recall the three hypotheses that motivate our work: \textbf{i)} Partner personas generation is plausible. \textbf{ii)} Generated personas are more diverse and interesting than the retrieved ground truth. \textbf{iii)} Generated personas boost succeeding dialogue generation. In the remaining section, we attempt to verify these claims and hypotheses.

\subsection{Main Results}
The main results are presented in Table \ref{main_result}.
\paragraph{Our Framework w/o RL} Even when trained without the reinforcement learning algorithm, our framework surpasses all our baselines and a comparison model. This phenomenon serves as an evidence for our claim \textbf{b)} and hypothesis \textbf{iii)}. Therefore, our framework more efficiently parameterises the training partner personas than our E2E baseline trained with ground truth partner personas. Also, our framework relaxes the constraint that requires the main and the auxiliary task to have a similar nature, which is unrealistic in our case.

\begin{table}[t!]
\small
\centering
\begin{tabular}{ccc}
\hline
\noalign{\vskip 1mm}  
\textbf{Model} & \textbf{Distinct-1}& \textbf{Distinct-2}\\
\noalign{\vskip 1mm}  
\hline
\hline
\noalign{\vskip 1mm}  
\multicolumn{3}{c}{\textbf{Validation Set}}\\
\noalign{\vskip 1mm}  
\hline
\noalign{\vskip 1mm}  
Ground Truth Label & $0.003120$ & $0.03575$  \\
\textbf{Our Generator} &  $\mathbf{0.003920}$ &  $\mathbf{0.05422}$  \\
\noalign{\vskip 1mm}  
\hline
\hline
\noalign{\vskip 1mm}  
\multicolumn{3}{c}{\textbf{Test Set}}\\
\noalign{\vskip 1mm}  
\hline
\noalign{\vskip 1mm}  
Ground Truth Label & 0.003267 & 0.03856  \\
\textbf{Our Generator} &  $\mathbf{0.004090}$ &  $\mathbf{0.05597}$  \\
\noalign{\vskip 1mm}  
\hline
\hline
\end{tabular}
\caption{\label{diversity_ppg}
Diversity evaluation for the generated partner personas against the ground truth partner personas.
}
\end{table}
\begin{table}[t!]
\small
\centering
\begin{tabular}{ccc}
\hline
\noalign{\vskip 1mm}  
\textbf{Model} & \textbf{Distinct-1}& \textbf{Distinct-2}\\
\noalign{\vskip 1mm}  
\hline
\hline
\noalign{\vskip 1mm}  
E2E w/ Full &  $0.01370$ & $0.1544$  \\
\textbf{Our Framework} & $\mathbf{0.01495}$ &  $\mathbf{0.1726}$  \\
\noalign{\vskip 1mm}  
\hline
\hline
\end{tabular}
\caption{\label{E2E-full}
Testing diversity evaluation for dialogue generation on our framework against the competitive E2E baseline with ground truth partner personas. We leave corresponding validation performance in Appendix \ref{valid}.
}
\end{table}

\paragraph{Our Framework w/ RL PPG\&DRG} Our proposed framework performs the best when trained with reinforcement learning. Again, this phenomenon strengthens our claim \textbf{b)} and hypothesis \textbf{iii)}. In addition, this phenomenon serves as direct evidence for our claim \textbf{d)}. This phenomenon implicitly verifies our hypothesis \textbf{i)}, because we train our critic network using ground truth partner personas. Since the reinforcement learning algorithm is effective, our critic network produces valid rewards, which indicates that the generated partner personas are reasonable. Furthermore, we present the change of validation performance for dialogue generation and partner personas generation in Figure \ref{3}.\footnote{The result is scaled for the sake of space and clarity.} We observe that both performances improve during the RL stage. Therefore, the improvement in partner personas generation intimately relates to the improvement in the succeeding dialogue generation; see our claim \textbf{b)} and \textbf{iii)}.
\begin{table*}[t!]
\small
\centering
\begin{tabular}{cccc}
\begin{tabularx}{\textwidth}{XXXX}
\hline
\noalign{\vskip 1mm}  
\textbf{Dialogue Context} &  \textbf{Generated Personas} & \textbf{Ours} & \textbf{E2E w/ Full} \\
\noalign{\vskip 1mm}  
\hline
\hline
\noalign{\vskip 1mm}    
Hi, want to go shopping with me? & I love to shop. I drive BMW. & Sure! I love shopping. I am a bit envious. & I am a student at a small community college.\\
\noalign{\vskip 1mm}  
\hline
\hline \noalign{\vskip 1mm}  
I am just happy that the apartment is quiet for once. &I live in an apartment.&I live in an apartment. I like to watch TV. &I live in California.\\
\noalign{\vskip 1mm}  
\hline
\hline \noalign{\vskip 1mm}  
I just went hiking also tired.&I love the outdoors.&I am sorry to hear that. Do you like to travel?&I do not know. I am not very good at it. \\
\noalign{\vskip 1mm}  
\hline
\end{tabularx}
\end{tabular}
\caption{\label{9}
A case study that compares our framework against the baseline that conditions on the ground truth partner personas. We denote the latter one as E2E w/ Full. For the sake of clarity and space, we present the preceding partner utterance as dialogue context, and we give the most salient generated partner personas.
}
\end{table*}
\begin{table}[t!]
\small
\centering
\begin{tabular}{ccc}
\hline
\noalign{\vskip 1mm}  
\textbf{Criteria} & \textbf{E2E w/ Full} & \textbf{Ours}\\
\noalign{\vskip 1mm}  
\hline
\hline
\noalign{\vskip 1mm}  
Appropriateness & \colorbox{lightgray}{$20$} & \colorbox{cyan}{\textcolor{white}{$\mathbf{80}$}}$^{\ddag}$ \\
\noalign{\vskip 1mm} 
Informativeness & \colorbox{lightgray}{31} & \colorbox{cyan}{\textcolor{white}{$\mathbf{69}$}}$^{\ddag}$   \\
\noalign{\vskip 1mm} 
Engagingness & \colorbox{lightgray}{$22$} & \colorbox{cyan}{\textcolor{white}{$\mathbf{78}$}}$^{\ddag}$   \\
\noalign{\vskip 1mm} 
Human-likeness & \colorbox{lightgray}{$24$} & \colorbox{cyan}{\textcolor{white}{$\mathbf{76}$}}$^{\ddag}$   \\
\noalign{\vskip 1mm}  
\hline
\hline
\end{tabular}
\caption{\label{human}
Human evaluation results in winning percentages. $\ddag$ indicates the results as passing a two-tailed binomial significance test with $p < 0.001$.
}
\end{table}
\paragraph{End-to-end (E2E) Baseline Models} Our E2E baseline with training partner personas has a better perplexity and worse extrinsic scores than the E2E baseline without partner personas. This might be due to the training-inference discrepancy, which could significantly impact the extrinsic evaluations.  \\

\paragraph{Multi-task Learning Comparison Model} Our multi-task learning comparison model produces inferior results. This is, however, predictable. First of all, the prior work \citep{MTML} constrained itself to a meta-learning framework. More concretely, the nature of partner personas generation and dialogue response generation largely differs. The output format of partner personas always initiates with first-person sentence starters, while dialogue responses give more general responses ranging from greetings to goodbyes.
\subsection{Case Study on Dialogue Response Generation}
\label{case}
Table \ref{9} depicts the case study for response generation. In the first case, our partner personas generator successfully gave a reasonable imagination that a person who likes shopping could be rich and drive a luxury car, which is not in the ground truth personas. This follows a surprising response `I am a bit envious'. In the second case, our personas generator successfully identifies that the partner lives in an apartment. Succeedingly, the response generator gives a relevant response and reduces the undesired hallucination\footnote{`I like to watch TV' is in the self personas, but `I live in California' is not. The latter one is thus a hallucination.} from the baseline model. In the third case, our partner personas generator generates a partner persona `I love the outdoors', which is not even in the ground truth personas. After that, the response generator produces a relevant response which also expresses empathy. 
These facts support our underlying hypotheses \textbf{i)}, \textbf{ii)}, and \textbf{iii)}. Furthermore, as in Table \ref{E2E-full}, we compare our proposed framework with an end-to-end dialogue agent with both training and inference ground truth partner personas. For these cherry-picked examples, our framework generates more informative and engaging responses than this competitive baseline. This verifies our claim \textbf{c)} and hypothesis \textbf{iii)}.

\begin{table*}[t!]
\small
\centering
\begin{tabular}{cc}
\begin{tabularx}{\textwidth}{XX}
\hline
\noalign{\vskip 1mm}  
\textbf{Generated Partner Personas} &  \textbf{Ground Truth Partner Personas} \\
\noalign{\vskip 1mm}  
\hline
\hline
\noalign{\vskip 1mm}    
I am working on a biology degree. I love book. 1984 is my favourite book. I am in college. & I am a student. I attend university and study biology. I am very studious and do not like to party or drink. I want to be marine biologist. \\
\noalign{\vskip 1mm}  
\hline
\hline \noalign{\vskip 1mm}  
I have been married for 6 years. I am a financial analyst for a brewery. I like to go to the casino on weekends. & I am a carpenter. I like playing poker. I do not have many friends. I have a wife and three kids. \\
\noalign{\vskip 1mm}  
\hline
\hline \noalign{\vskip 1mm}  
I like to watch football. My friends like watching it too. Its great fun. We drink beer and eat food. & I love watching football on Sunday. I have three dogs. My favroutie food is cheese piazza. I am a hair stylist.\\ 
\noalign{\vskip 1mm}  
\hline
\end{tabularx}
\end{tabular}
\caption{\label{ppg-examples}
A case study to show that our generated personas are relevant, informative and coherent. For the sake of space, we present more cases in Appendix \ref{appcase}.
}
\end{table*}

\subsection{Human Evaluation}
We hired experienced annotators who have degrees relevant to English Linguistics or Applied Linguistics. We present a questionnaire composed of 280 questions with randomly sampled 70 testing instances. Three annotators compare model outputs in an A/B setting. As in previous work \citep{2021arXiv210904084Z} and ACUTE-Evals \citep{2019arXiv190903087L}, annotators follow the criteria:
\begin{itemize}
    \item \textbf{(Appropriateness)}: \textit{"Who is more appropriate given the previous dialogue context?"}
    \item \textbf{(Informativeness)}: \textit{"Who is more diverse instead of null answers such as I do not know?"}
    \item \textbf{(Engagingness)}: \textit{"Who would you prefer to talk with for a long conversation?"}
    \item \textbf{(Human-likeness)}: \textit{"Which speaker do you think sounds more like a real person?"}
    
\end{itemize}
Table \ref{human} presents the human evaluation results. It is exciting to see our framework trained under reinforcement learning surpasses the end-to-end model that leverages both training and inference ground truth partner personas, from all the aspects, significantly. This supports our claim \textbf{c)} and \textbf{d)}.

\subsection{Ablation Study on Reinforcement Learning}
\label{ablation}
Table \ref{main_result} presents an ablation study on the framework when only one of the modules, namely partner personas generator (PPG) or dialogue response generator (DRG), is trained under reinforcement learning. Our framework exceeds these two variants in all the metrics except for perplexity, which aligns with the prior work \citep{2020arXiv200413637R}.\footnote{PPL does not always correlate well with other metrics.}

\subsection{Case Study on Partner Personas Generation}
As depicted in Table \ref{diversity_ppg}, we observe that our partner personas generator generates more diverse partner personas compared to the ground truth partner personas label, which is essentially the upper bound for retrieval-based partner personas predictor. This phenomenon verifies our claim \textbf{a)} and hypothesis \textbf{i)}, indicating our generator produces even more informative and interesting partner personas than the ground truth partner personas.
\par
As depicted in Table \ref{ppg-examples}, our partner personas generator can generate plausible partner personas which are relevant to the ground truth partner personas. Our partner personas generator can give fascinating but reasonable imagination that is not even in the dialogue context or the ground truth partner personas. In the first case, the generator successfully identified the partner as a student studying biology. The generator recognizes the partner as being married in the second case, which is not even mentioned in the dialogue context. This phenomenon could be a matter of the fact that personas could be semantically closer to each other when they frequently co-occur in the training set. In the third case, the generator generates diverse personas, saying that the partner would drink beer and eat food while watching football, which is even not in any of the dialogue context or the ground truth partner personas. This verifies claim \textbf{a)} and hypothesis \textbf{i)} and \textbf{ii)}, We depict more case studies in Appendix \ref{appcase} to show that our personas generator generates informative and coherent partner personas.\footnote{Our partner personas generator is even capable of producing unseen personas. An offensiveness check thus is necessary for the actual usage, as in prior works \citep{2021arXiv210811830B}.}

\section{Conclusion}
Our novel framework incorporates partner personas generation into succeeding dialogue response generation. First of all, our proposed framework mitigates the cold-start problem in practical applications when ground truth partner personas could be missing during inference. The experimental results with both automatic and human evaluation demonstrate that our framework generates informative and coherent partner personas, even compared to the \textit{ground truth} partner personas, yet still reasonable and relevant. This enhances the succeeding response generation, thus surpassing our baselines and producing responses that are more diverse and engaging than our baseline conditioned on the \textit{ground truth} partner personas. We employ reinforcement learning with a dedicatedly designed critic network that boosts our framework. Extensive case studies demonstrate that our framework can generate satisfying dialogue responses and partner personas. Finally, our framework gives better explainability and reduces the demands for external databases for partner personas.
\bibliography{anthology,custom}

\begin{thebibliography}{36}
\expandafter\ifx\csname natexlab\endcsname\relax\def\natexlab#1{#1}\fi

\bibitem[{{Baheti} et~al.(2021){Baheti}, {Sap}, {Ritter}, and
  {Riedl}}]{2021arXiv210811830B}
Ashutosh {Baheti}, Maarten {Sap}, Alan {Ritter}, and Mark {Riedl}. 2021.
\newblock \href {http://arxiv.org/abs/2108.11830} {{Just Say No: Analyzing the
  Stance of Neural Dialogue Generation in Offensive Contexts}}.
\newblock \emph{arXiv e-prints}, page arXiv:2108.11830.

\bibitem[{Banerjee and Lavie(2005)}]{banerjee-lavie-2005-meteor2}
Satanjeev Banerjee and Alon Lavie. 2005.
\newblock \href {https://aclanthology.org/W05-0909} {{METEOR}: An automatic
  metric for {MT} evaluation with improved correlation with human judgments}.
\newblock In \emph{Proceedings of the {ACL} Workshop on Intrinsic and Extrinsic
  Evaluation Measures for Machine Translation and/or Summarization}, pages
  65--72, Ann Arbor, Michigan. Association for Computational Linguistics.

\bibitem[{{Bao} et~al.(2021){Bao}, {He}, {Wang}, {Wu}, {Wang}, {Wu}, {Wu},
  {Guo}, {Lu}, {Huang}, {Tian}, {Xu}, {Lin}, and {Niu}}]{2021arXiv210909519B}
Siqi {Bao}, Huang {He}, Fan {Wang}, Hua {Wu}, Haifeng {Wang}, Wenquan {Wu},
  Zhihua {Wu}, Zhen {Guo}, Hua {Lu}, Xinxian {Huang}, Xin {Tian}, Xinchao {Xu},
  Yingzhan {Lin}, and Zhengyu {Niu}. 2021.
\newblock \href {http://arxiv.org/abs/2109.09519} {{PLATO-XL: Exploring the
  Large-scale Pre-training of Dialogue Generation}}.
\newblock \emph{arXiv e-prints}, page arXiv:2109.09519.

\bibitem[{Cai et~al.(2019)Cai, Wang, Bi, Tu, Liu, Lam, and
  Shi}]{cai-etal-2019-skeleton}
Deng Cai, Yan Wang, Wei Bi, Zhaopeng Tu, Xiaojiang Liu, Wai Lam, and Shuming
  Shi. 2019.
\newblock \href {https://doi.org/10.18653/v1/N19-1124} {Skeleton-to-response:
  Dialogue generation guided by retrieval memory}.
\newblock In \emph{Proceedings of the 2019 Conference of the North {A}merican
  Chapter of the Association for Computational Linguistics: Human Language
  Technologies, Volume 1 (Long and Short Papers)}, pages 1219--1228,
  Minneapolis, Minnesota. Association for Computational Linguistics.

\bibitem[{{Deng} et~al.(2021){Deng}, {Li}, {Sun}, {Ding}, and
  {Lam}}]{2021arXiv210509710D}
Yang {Deng}, Yaliang {Li}, Fei {Sun}, Bolin {Ding}, and Wai {Lam}. 2021.
\newblock \href {http://arxiv.org/abs/2105.09710} {{Unified Conversational
  Recommendation Policy Learning via Graph-based Reinforcement Learning}}.
\newblock \emph{arXiv e-prints}, page arXiv:2105.09710.

\bibitem[{Gu et~al.(2019)Gu, Ling, Zhu, and Liu}]{DIM}
Jia-Chen Gu, Zhen-Hua Ling, Xiaodan Zhu, and Quan Liu. 2019.
\newblock \href {https://doi.org/10.18653/v1/D19-1193} {Dually interactive
  matching network for personalized response selection in retrieval-based
  chatbots}.
\newblock In \emph{Proceedings of the 2019 Conference on Empirical Methods in
  Natural Language Processing and the 9th International Joint Conference on
  Natural Language Processing (EMNLP-IJCNLP)}, pages 1845--1854, Hong Kong,
  China. Association for Computational Linguistics.

\bibitem[{{Gu} et~al.(2021){Gu}, {Liu}, {Ling}, {Liu}, {Chen}, and
  {Zhu}}]{PTMT}
Jia-Chen {Gu}, Hui {Liu}, Zhen-Hua {Ling}, Quan {Liu}, Zhigang {Chen}, and
  Xiaodan {Zhu}. 2021.
\newblock \href {http://arxiv.org/abs/2105.09050} {{Partner Matters! An
  Empirical Study on Fusing Personas for Personalized Response Selection in
  Retrieval-Based Chatbots}}.
\newblock \emph{arXiv e-prints}, page arXiv:2105.09050.

\bibitem[{Kim et~al.(2020)Kim, Kim, and Kim}]{kim-etal-2020-will}
Hyunwoo Kim, Byeongchang Kim, and Gunhee Kim. 2020.
\newblock \href {https://doi.org/10.18653/v1/2020.emnlp-main.65} {Will {I}
  sound like me? improving persona consistency in dialogues through pragmatic
  self-consciousness}.
\newblock In \emph{Proceedings of the 2020 Conference on Empirical Methods in
  Natural Language Processing (EMNLP)}, pages 904--916, Online. Association for
  Computational Linguistics.

\bibitem[{{Kingma} and {Ba}(2014)}]{adam}
Diederik~P. {Kingma} and Jimmy {Ba}. 2014.
\newblock \href {http://arxiv.org/abs/1412.6980} {{Adam: A Method for
  Stochastic Optimization}}.
\newblock \emph{arXiv e-prints}, page arXiv:1412.6980.

\bibitem[{{Lee} et~al.(2021){Lee}, {Aik Lee}, and {Gan}}]{MTML}
Jing~Yang {Lee}, Kong {Aik Lee}, and Woon~Seng {Gan}. 2021.
\newblock \href {http://arxiv.org/abs/2108.03377} {{Generating Personalized
  Dialogue via Multi-Task Meta-Learning}}.
\newblock \emph{arXiv e-prints}, page arXiv:2108.03377.

\bibitem[{{Li} et~al.(2018){Li}, {Huang}, {He}, {Zhang}, and
  {Sun}}]{2018arXiv180400861L}
Dianqi {Li}, Qiuyuan {Huang}, Xiaodong {He}, Lei {Zhang}, and Ming-Ting {Sun}.
  2018.
\newblock \href {http://arxiv.org/abs/1804.00861} {{Generating Diverse and
  Accurate Visual Captions by Comparative Adversarial Learning}}.
\newblock \emph{arXiv e-prints}, page arXiv:1804.00861.

\bibitem[{{Li} et~al.(2015){Li}, {Galley}, {Brockett}, {Gao}, and
  {Dolan}}]{2015arXiv151003055L}
Jiwei {Li}, Michel {Galley}, Chris {Brockett}, Jianfeng {Gao}, and Bill
  {Dolan}. 2015.
\newblock \href {http://arxiv.org/abs/1510.03055} {{A Diversity-Promoting
  Objective Function for Neural Conversation Models}}.
\newblock \emph{arXiv e-prints}, page arXiv:1510.03055.

\bibitem[{{Li} et~al.(2016){Li}, {Monroe}, {Ritter}, {Galley}, {Gao}, and
  {Jurafsky}}]{RLChat}
Jiwei {Li}, Will {Monroe}, Alan {Ritter}, Michel {Galley}, Jianfeng {Gao}, and
  Dan {Jurafsky}. 2016.
\newblock \href {http://arxiv.org/abs/1606.01541} {{Deep Reinforcement Learning
  for Dialogue Generation}}.
\newblock \emph{arXiv e-prints}, page arXiv:1606.01541.

\bibitem[{{Li} et~al.(2019){Li}, {Weston}, and {Roller}}]{2019arXiv190903087L}
Margaret {Li}, Jason {Weston}, and Stephen {Roller}. 2019.
\newblock \href {http://arxiv.org/abs/1909.03087} {{ACUTE-EVAL: Improved
  Dialogue Evaluation with Optimized Questions and Multi-turn Comparisons}}.
\newblock \emph{arXiv e-prints}, page arXiv:1909.03087.

\bibitem[{{Li} et~al.(2020){Li}, {Lei}, {Wu}, {He}, {Jiang}, and
  {Chua}}]{2020arXiv200512979L}
Shijun {Li}, Wenqiang {Lei}, Qingyun {Wu}, Xiangnan {He}, Peng {Jiang}, and
  Tat-Seng {Chua}. 2020.
\newblock \href {http://arxiv.org/abs/2005.12979} {{Seamlessly Unifying
  Attributes and Items: Conversational Recommendation for Cold-Start Users}}.
\newblock \emph{arXiv e-prints}, page arXiv:2005.12979.

\bibitem[{Lin(2004)}]{rouge}
Chin-Yew Lin. 2004.
\newblock \href {https://www.aclweb.org/anthology/W04-1013} {{ROUGE}: A package
  for automatic evaluation of summaries}.
\newblock In \emph{Text Summarization Branches Out}, pages 74--81, Barcelona,
  Spain. Association for Computational Linguistics.

\bibitem[{Madotto et~al.(2020)Madotto, Cahyawijaya, Winata, Xu, Liu, Lin, and
  Fung}]{madotto-etal-2020-learning}
Andrea Madotto, Samuel Cahyawijaya, Genta~Indra Winata, Yan Xu, Zihan Liu,
  Zhaojiang Lin, and Pascale Fung. 2020.
\newblock \href {https://doi.org/10.18653/v1/2020.findings-emnlp.215} {Learning
  knowledge bases with parameters for task-oriented dialogue systems}.
\newblock In \emph{Findings of the Association for Computational Linguistics:
  EMNLP 2020}, pages 2372--2394, Online. Association for Computational
  Linguistics.

\bibitem[{Mazar{\'e} et~al.(2018)Mazar{\'e}, Humeau, Raison, and Bordes}]{FTPC}
Pierre-Emmanuel Mazar{\'e}, Samuel Humeau, Martin Raison, and Antoine Bordes.
  2018.
\newblock \href {https://doi.org/10.18653/v1/D18-1298} {Training millions of
  personalized dialogue agents}.
\newblock In \emph{Proceedings of the 2018 Conference on Empirical Methods in
  Natural Language Processing}, pages 2775--2779, Brussels, Belgium.
  Association for Computational Linguistics.

\bibitem[{{Miller} et~al.(2017){Miller}, {Feng}, {Fisch}, {Lu}, {Batra},
  {Bordes}, {Parikh}, and {Weston}}]{2017arXiv170506476M}
Alexander~H. {Miller}, Will {Feng}, Adam {Fisch}, Jiasen {Lu}, Dhruv {Batra},
  Antoine {Bordes}, Devi {Parikh}, and Jason {Weston}. 2017.
\newblock \href {http://arxiv.org/abs/1705.06476} {{ParlAI: A Dialog Research
  Software Platform}}.
\newblock \emph{arXiv e-prints}, page arXiv:1705.06476.

\bibitem[{Radford et~al.(2019)Radford, Wu, Child, Luan, Amodei, and
  Sutskever}]{gpt2}
Alec Radford, Jeff Wu, Rewon Child, David Luan, Dario Amodei, and Ilya
  Sutskever. 2019.
\newblock Language models are unsupervised multitask learners.

\bibitem[{{Roller} et~al.(2020){Roller}, {Dinan}, {Goyal}, {Ju}, {Williamson},
  {Liu}, {Xu}, {Ott}, {Shuster}, {Smith}, {Boureau}, and
  {Weston}}]{2020arXiv200413637R}
Stephen {Roller}, Emily {Dinan}, Naman {Goyal}, Da~{Ju}, Mary {Williamson},
  Yinhan {Liu}, Jing {Xu}, Myle {Ott}, Kurt {Shuster}, Eric~M. {Smith}, Y-Lan
  {Boureau}, and Jason {Weston}. 2020.
\newblock \href {http://arxiv.org/abs/2004.13637} {{Recipes for building an
  open-domain chatbot}}.
\newblock \emph{arXiv e-prints}, page arXiv:2004.13637.

\bibitem[{Roman~Roman et~al.(2020)Roman~Roman, Bisk, Thomason, Celikyilmaz, and
  Gao}]{RMM}
Homero Roman~Roman, Yonatan Bisk, Jesse Thomason, Asli Celikyilmaz, and
  Jianfeng Gao. 2020.
\newblock \href {https://doi.org/10.18653/v1/2020.findings-emnlp.157} {{RMM}: A
  recursive mental model for dialogue navigation}.
\newblock In \emph{Findings of the Association for Computational Linguistics:
  EMNLP 2020}, pages 1732--1745, Online. Association for Computational
  Linguistics.

\bibitem[{{Saleh} et~al.(2019){Saleh}, {Jaques}, {Ghandeharioun}, {Hanwen
  Shen}, and {Picard}}]{2019arXiv190907547S}
Abdelrhman {Saleh}, Natasha {Jaques}, Asma {Ghandeharioun}, Judy {Hanwen Shen},
  and Rosalind {Picard}. 2019.
\newblock \href {http://arxiv.org/abs/1909.07547} {{Hierarchical Reinforcement
  Learning for Open-Domain Dialog}}.
\newblock \emph{arXiv e-prints}, page arXiv:1909.07547.

\bibitem[{{Sanh} et~al.(2019){Sanh}, {Debut}, {Chaumond}, and
  {Wolf}}]{2019arXiv191001108S}
Victor {Sanh}, Lysandre {Debut}, Julien {Chaumond}, and Thomas {Wolf}. 2019.
\newblock \href {http://arxiv.org/abs/1910.01108} {{DistilBERT, a distilled
  version of BERT: smaller, faster, cheaper and lighter}}.
\newblock \emph{arXiv e-prints}, page arXiv:1910.01108.

\bibitem[{Song et~al.(2021)Song, Wang, Zhang, Zhang, and
  Liu}]{song-etal-2021-bob}
Haoyu Song, Yan Wang, Kaiyan Zhang, Wei-Nan Zhang, and Ting Liu. 2021.
\newblock \href {https://doi.org/10.18653/v1/2021.acl-long.14} {{B}o{B}: {BERT}
  over {BERT} for training persona-based dialogue models from limited
  personalized data}.
\newblock In \emph{Proceedings of the 59th Annual Meeting of the Association
  for Computational Linguistics and the 11th International Joint Conference on
  Natural Language Processing (Volume 1: Long Papers)}, pages 167--177, Online.
  Association for Computational Linguistics.

\bibitem[{Song et~al.(2020)Song, Wang, Zhang, Liu, and
  Liu}]{song-etal-2020-generate}
Haoyu Song, Yan Wang, Wei-Nan Zhang, Xiaojiang Liu, and Ting Liu. 2020.
\newblock \href {https://doi.org/10.18653/v1/2020.acl-main.516} {Generate,
  delete and rewrite: A three-stage framework for improving persona consistency
  of dialogue generation}.
\newblock In \emph{Proceedings of the 58th Annual Meeting of the Association
  for Computational Linguistics}, pages 5821--5831, Online. Association for
  Computational Linguistics.

\bibitem[{{Song} et~al.(2019){Song}, {Zhang}, {Cui}, {Wang}, and
  {Liu}}]{2019arXiv190512188S}
Haoyu {Song}, Wei-Nan {Zhang}, Yiming {Cui}, Dong {Wang}, and Ting {Liu}. 2019.
\newblock \href {http://arxiv.org/abs/1905.12188} {{Exploiting Persona
  Information for Diverse Generation of Conversational Responses}}.
\newblock \emph{arXiv e-prints}, page arXiv:1905.12188.

\bibitem[{{Sutskever} et~al.(2014){Sutskever}, {Vinyals}, and
  {Le}}]{2014arXiv1409.3215S}
Ilya {Sutskever}, Oriol {Vinyals}, and Quoc~V. {Le}. 2014.
\newblock \href {http://arxiv.org/abs/1409.3215} {{Sequence to Sequence
  Learning with Neural Networks}}.
\newblock \emph{arXiv e-prints}, page arXiv:1409.3215.

\bibitem[{Williams(1992)}]{policygradients}
Ronald~J. Williams. 1992.
\newblock \href {https://doi.org/10.1007/BF00992696} {Simple statistical
  gradient-following algorithms for connectionist reinforcement learning}.
\newblock \emph{Mach. Learn.}, 8(3–4):229–256.

\bibitem[{Wolf et~al.(2020)Wolf, Debut, Sanh, Chaumond, Delangue, Moi, Cistac,
  Rault, Louf, Funtowicz, Davison, Shleifer, von Platen, Ma, Jernite, Plu, Xu,
  Scao, Gugger, Drame, Lhoest, and Rush}]{wolf-etal-2020-transformers}
Thomas Wolf, Lysandre Debut, Victor Sanh, Julien Chaumond, Clement Delangue,
  Anthony Moi, Pierric Cistac, Tim Rault, Rémi Louf, Morgan Funtowicz, Joe
  Davison, Sam Shleifer, Patrick von Platen, Clara Ma, Yacine Jernite, Julien
  Plu, Canwen Xu, Teven~Le Scao, Sylvain Gugger, Mariama Drame, Quentin Lhoest,
  and Alexander~M. Rush. 2020.
\newblock \href {https://www.aclweb.org/anthology/2020.emnlp-demos.6}
  {Transformers: State-of-the-art natural language processing}.
\newblock In \emph{Proceedings of the 2020 Conference on Empirical Methods in
  Natural Language Processing: System Demonstrations}, pages 38--45, Online.
  Association for Computational Linguistics.

\bibitem[{{Wolf} et~al.(2019){Wolf}, {Sanh}, {Chaumond}, and
  {Delangue}}]{2019arXiv190108149W}
Thomas {Wolf}, Victor {Sanh}, Julien {Chaumond}, and Clement {Delangue}. 2019.
\newblock \href {http://arxiv.org/abs/1901.08149} {{TransferTransfo: A Transfer
  Learning Approach for Neural Network Based Conversational Agents}}.
\newblock \emph{arXiv e-prints}, page arXiv:1901.08149.

\bibitem[{{Xu} et~al.(2021){Xu}, {Ishii}, {Liu}, {Indra Winata}, {Su},
  {Madotto}, and {Fung}}]{free}
Yan {Xu}, Etsuko {Ishii}, Zihan {Liu}, Genta {Indra Winata}, Dan {Su}, Andrea
  {Madotto}, and Pascale {Fung}. 2021.
\newblock \href {http://arxiv.org/abs/2105.06232} {{Retrieval-Free
  Knowledge-Grounded Dialogue Response Generation with Adapters}}.
\newblock \emph{arXiv e-prints}, page arXiv:2105.06232.

\bibitem[{Zhang et~al.(2018)Zhang, Dinan, Urbanek, Szlam, Kiela, and
  Weston}]{PERSONACHAT}
Saizheng Zhang, Emily Dinan, Jack Urbanek, Arthur Szlam, Douwe Kiela, and Jason
  Weston. 2018.
\newblock \href {https://doi.org/10.18653/v1/P18-1205} {Personalizing dialogue
  agents: {I} have a dog, do you have pets too?}
\newblock In \emph{Proceedings of the 56th Annual Meeting of the Association
  for Computational Linguistics (Volume 1: Long Papers)}, pages 2204--2213,
  Melbourne, Australia. Association for Computational Linguistics.

\bibitem[{{Zhang} et~al.(2019){Zhang}, {Sun}, {Galley}, {Chen}, {Brockett},
  {Gao}, {Gao}, {Liu}, and {Dolan}}]{2019arXiv191100536Z}
Yizhe {Zhang}, Siqi {Sun}, Michel {Galley}, Yen-Chun {Chen}, Chris {Brockett},
  Xiang {Gao}, Jianfeng {Gao}, Jingjing {Liu}, and Bill {Dolan}. 2019.
\newblock \href {http://arxiv.org/abs/1911.00536} {{DialoGPT: Large-Scale
  Generative Pre-training for Conversational Response Generation}}.
\newblock \emph{arXiv e-prints}, page arXiv:1911.00536.

\bibitem[{{Zhao} et~al.(2019){Zhao}, {Tao}, {Wu}, {Xu}, {Zhao}, and
  {Yan}}]{DGMN}
Xueliang {Zhao}, Chongyang {Tao}, Wei {Wu}, Can {Xu}, Dongyan {Zhao}, and Rui
  {Yan}. 2019.
\newblock \href {http://arxiv.org/abs/1906.04362} {{A Document-grounded
  Matching Network for Response Selection in Retrieval-based Chatbots}}.
\newblock \emph{arXiv e-prints}, page arXiv:1906.04362.

\bibitem[{{Zou} et~al.(2021){Zou}, {Liu}, {Hu}, and
  {Zhang}}]{2021arXiv210904084Z}
Yicheng {Zou}, Zhihua {Liu}, Xingwu {Hu}, and Qi~{Zhang}. 2021.
\newblock \href {http://arxiv.org/abs/2109.04084} {{Thinking Clearly, Talking
  Fast: Concept-Guided Non-Autoregressive Generation for Open-Domain Dialogue
  Systems}}.
\newblock \emph{arXiv e-prints}, page arXiv:2109.04084.

\end{thebibliography}
\bibliographystyle{acl_natbib}
\appendix
\section{Validation Performance}
\label{valid}
Table \ref{valid_result} presents the corresponding validation performance to Table \ref{main_result}. Table \ref{E2E-full-val} presents the corresponding validation performance to Table \ref{E2E-full}.
\begin{table}[h!]
\small
\centering
\begin{tabular}{ccc}
\hline
\noalign{\vskip 1mm}  
\textbf{Model} & \textbf{Distinct-1}& \textbf{Distinct-2}\\
\noalign{\vskip 1mm}  
\hline
\hline
\noalign{\vskip 1mm}  
E2E w/ Full & $0.01293$ & $0.1484$  \\
\textbf{Our Framework} &  $\mathbf{0.01339}$ &  $\mathbf{0.1591}$  \\
\noalign{\vskip 1mm}  
\hline
\hline
\end{tabular}
\caption{\label{E2E-full-val}
Validation diversity evaluation for dialogue generation that compares our framework against the competitive E2E baseline with ground truth partner personas, as promised in Table \ref{E2E-full}.
}
\end{table}
\section{More Case Studies}
\label{appcase}
Table \ref{moreppg-examples} presents extensive case studies for partner personas generation. These examples indicate that our framework can generate informative and coherent partner personas. We highlight in pink for informativeness and in yellow for coherence.
\begin{table*}
\small
\centering
\begin{tabular}{lccccc}
\hline
\noalign{\vskip 1mm}  
\textbf{Model} & \textbf{PPL$\downarrow$} & \textbf{ROUGE} & \textbf{METEOR} & \textbf{Distinct-1}& \textbf{Distinct-2}\\
\noalign{\vskip 1mm}  
\hline
\hline
\noalign{\vskip 1mm}  
E2E w/o Partner Personas & $20.02$ & $16.42$ & $24.46$ & $0.01168$  & $0.1422$ \\
E2E w/ Training Partner Personas & $19.71$ & $16.27$ & $24.10$ & $0.01221$  & $0.1413$ \\
Multi-task Learning \citep{MTML} & $218.7$ & $10.27$ & $12.80$ & $0.00395$  & $0.0560$ \\
\noalign{\vskip 1mm}  
\hline
\hline
\noalign{\vskip 1mm}  
Our Framework w/ RL PPG & $18.31$ & $\mathbf{16.77}$ & $25.20$ & $0.01329$  & $0.1511$ \\
Our Framework w/ RL DRG & $18.70$ & $16.69$ & $25.64$ & $\mathbf{0.01345}$  & $0.1575$ \\
\noalign{\vskip 1mm} 
\hline
\noalign{\vskip 1mm} 
Our Framework w/o RL & $18.56$ & $16.71$ & $24.79$ & $0.01322$  & $0.1490$ \\
\textbf{Our Framework w/ RL PPG\&DRG} & $\mathbf{18.19}$ &  $\mathbf{16.77}$ &  $\mathbf{26.14}$ &  $0.01339$  &  $\mathbf{0.1591}$ \\ 
\noalign{\vskip 1mm}  
\hline
\hline
\end{tabular}
\caption{\label{valid_result}
Corresponding validation performance as promised in Table \ref{main_result}.\
}
\end{table*}

\begin{table*}[t!]
\small
\centering
\begin{tabular}{c}
\begin{tabularx}{\textwidth}{X}
\hline
\noalign{\vskip 1mm}  
\textbf{Generated Partner Personas} \\
\noalign{\vskip 1mm}  
\hline
\hline
\noalign{\vskip 1mm}    
\textbf{Personas A}: I am a \hl{shy person}, but I love to sing. Until recently, I ve never been able to sing in front of anyone. Anyways, I decided to give it a try and participaed in an \colorbox{pink}{audition for a talent show}. \hl{My shyness} made me panick and I didn t show up.  \\
\noalign{\vskip 1mm}  
\hline
\hline \noalign{\vskip 1mm}  
\textbf{Personas B}: I play the violin. I am \hl{married} with 5 kids. I am nurse. I met my \hl{husband} when I was \colorbox{pink}{a freshman in college}. \\
\noalign{\vskip 1mm}  
\hline
\hline \noalign{\vskip 1mm}  
\textbf{Personas C}: I am a \hl{soccer player}. I am a goalie. \hl{My number} is 42. \colorbox{pink}{Nike cleats} are my favorite. I joined a \hl{new team} last month.   \\
\noalign{\vskip 1mm}  
\hline
\hline \noalign{\vskip 1mm}  
\textbf{Personas D}: I have \hl{two kids}, ages 2 and 6. I am from \colorbox{pink}{sterling heights, michigan}. My favorite movie is \colorbox{pink}{titanic}. I work part time at aldis. My \hl{husband} owns a small auto repair shop.  \\
\noalign{\vskip 1mm}  
\hline
\hline \noalign{\vskip 1mm}  
\textbf{Personas E}: I am a \hl{retired computer programmer}. I have one grandson and one daughter. I just \hl{turned 77}. I love animals. I like watching \colorbox{pink}{british tv shows and movies}. \\
\noalign{\vskip 1mm}  
\hline
\hline \noalign{\vskip 1mm}  
\textbf{Personas F}: I like to go \hl{hunting}. I like to remodel homes. I like to \hl{shoot a bow}. My favorite holiday is \colorbox{pink}{halloween}. I like to go shopping with my daughters.  \\
\noalign{\vskip 1mm}  
\hline
\hline \noalign{\vskip 1mm}  
\textbf{Personas G}: I have a \hl{large cd collection}. I collect stamps. Favorite \hl{band} is the \colorbox{pink}{beetles}. I play the bass. I like \colorbox{pink}{vintage furniture}.  \\
\noalign{\vskip 1mm}  
\hline
\hline \noalign{\vskip 1mm}  
\textbf{Personas H}: I have a \hl{daughter}. I am a \colorbox{pink}{yoga instructor}. I enjoy shopping. I have two \hl{adopted kids}. \\
\noalign{\vskip 1mm}  
\hline
\hline \noalign{\vskip 1mm}  
\textbf{Personas I}: I like to drink wine. I enjoy reading \hl{history books}. I am a teacher. I love to \hl{write stories} while \colorbox{pink}{sitting in the grass} in my back yard. I grew up in new hampshire. \\
\noalign{\vskip 1mm}  
\hline
\hline \noalign{\vskip 1mm}  
\textbf{Personas J}: I am \hl{retired}. I stay active. I have \hl{eight grandchildren}. I have good health. \\
\noalign{\vskip 1mm}  
\hline
\hline \noalign{\vskip 1mm}  
\textbf{Personas K}: I m a \hl{student}. I like to go out to eat. I like listening to other \hl{rap music} too. One of my \colorbox{pink}{favorite artists is drake}. A hobby of mine is the \hl{drums}. I also enjoy cooking . \\
\noalign{\vskip 1mm}  
\hline
\hline \noalign{\vskip 1mm}  
\textbf{Personas L}: I have two children. I like to go on walks. I am from mexico. I used to be a \hl{chef}, but I am a teacher now. I like to \hl{bake}.  \\
\noalign{\vskip 1mm}  
\hline
\hline \noalign{\vskip 1mm}  
\textbf{Personas M}: I drive a \colorbox{pink}{prius}. My mom stays at home. I was \hl{adopted} when I was a baby. My \hl{adopted dad} works at hp. \\
\noalign{\vskip 1mm}  
\hline
\hline \noalign{\vskip 1mm}  
\textbf{Personas N}: \hl{I love youtube}. My father works in \colorbox{pink}{advertising agency}. \hl{I have my own channel}. I enjoy making let s plays. \\
\noalign{\vskip 1mm}  
\hline
\hline \noalign{\vskip 1mm}  
\textbf{Personas O}: I \hl{like to cook}. I \hl{am a foodie}. I love to chat with my friends. I \hl{love kids and dogs}. I like to \hl{go shopping with my daughters}. \\
\noalign{\vskip 1mm}  
\hline
\hline \noalign{\vskip 1mm}  
\textbf{Personas P}: My family hates my \hl{fiance}. We will be traveling to \colorbox{pink}{niagra falls} \hl{for our honeymoon. We are getting married in a park}. My dog is the ring bearer. \\
\noalign{\vskip 1mm}  
\hline
\end{tabularx}
\end{tabular}
\caption{\label{moreppg-examples}
Case studies as promised in Table \ref{ppg-examples}. We highlight in pink for informativeness and in yellow for coherence.
}
\end{table*}
\end{document}